\documentclass[runningheads]{llncs}
\usepackage{booktabs} 
\usepackage{color}
\usepackage{amsfonts}
\usepackage{graphicx}
\usepackage{caption}
\usepackage{algorithm,algorithmic}
\usepackage{amsmath}
\usepackage{graphics}
\usepackage{multirow}
\usepackage{bm}
\usepackage[mathscr]{euscript}
\usepackage{setspace}
\usepackage{geometry}
\begin{document}

\title{MAD-GAN: Multivariate Anomaly Detection for Time Series Data with Generative Adversarial Networks}
%
%
\author{
Dan Li\inst{1}\orcidID{0000-1111-2222-3333} \and
Dacheng Chen\inst{2,3}\orcidID{1111-2222-3333-4444} \and
Lei Shi\inst{3}\orcidID{2222--3333-4444-5555} \and
Baihong Jin\inst{3}\orcidID{2222--3333-4444-5555} \and
Jonathan Goh\inst{3}\orcidID{2222--3333-4444-5555} \and
See-Kiong Ng\inst{3}\orcidID{2222--3333-4444-5555} \and}
\author{Dan Li\inst{1} \and
Dacheng Chen\inst{1} \and
Lei Shi\inst{1} \and
Baihong Jin\inst{2} \and
Jonathan Goh\inst{3} \and
See-Kiong Ng\inst{1}}

\authorrunning{D. Li et al.}

\institute{
Institute of Data Science, National University of Singapore, 3 Research Link Singapore 117602\\
\and
Department of Electrical Engineering and Computer Sciences, University of California, Berkeley, CA 94720 USA\\
\and
ST Electronics (Info Security) Pte Ltd, 100 Jurong East Street 21 Singapore 609602 \\
\thanks{The code is available at https://github.com/LiDan456/MAD-GANs}
}

\date{}
\maketitle              
\begin{abstract}

The prevalence of networked sensors and actuators in many real-world systems such as smart buildings, factories, power plants, and data centres generate substantial amounts of multivariate time series data for these systems. Many of these cyber-physical systems (CPSs) are engineered for mission-critical tasks and are thus targets for cyber-attacks. The rich sensor data can be continuously monitored for intrusion events through anomaly detection. However, conventional threshold-based anomaly detection methods are inadequate due to the dynamic complexities of these systems, while supervised machine learning methods are unable to exploit the large amounts of data due to the lack of labeled data. On the other hand, current unsupervised machine learning approaches have not fully exploited the spatial-temporal correlation and other dependencies amongst the multiple variables (sensors/actuators) in the system for detecting anomalies. Most of the current techniques also employed simple comparison between the present states and predicted normal ranges for anomaly detection, which can be inadequate given the  highly dynamic behaviors of the systems.  In this work, we propose an unsupervised multivariate anomaly detection method based on Generative Adversarial Networks (GANs), using the Long-Short-Term-Memory Recurrent Neural Networks (LSTM-RNN) as the base models (namely, the generator and discriminator) in the GAN framework to capture the temporal correlation of time series distributions.  Instead of treating each data stream independently, our proposed \underline{M}ultivariate \underline{A}nomaly \underline{D}etection with GAN (MAD-GAN) framework considers the entire variable set concurrently to capture the latent interactions amongst the variables. We also fully exploit both the generator and discriminator produced by the GAN, using a novel anomaly score called DR-score to detect anomalies by discrimination and reconstruction.   We have tested our proposed MAD-GAN using two recent datasets collected from real world CPS: the Secure Water Treatment (SWaT) and the Water Distribution (WADI) datasets. Our experimental results showed that the proposed MAD-GAN is effective in reporting anomalies caused by various cyber-intrusions compared in these complex real-world systems.

\end{abstract}
\section{Introduction}
\label{sec:Intro}

Today's Cyber-Physical Systems (CPSs) such as smart buildings, factories, power plants, and data centres are large, complex, and affixed with networked sensors and actuators that generate substantial amounts of multivariate time series data that can be used to continuously monitor the CPS' working conditions to detect anomalies in time \cite{Jonathan2017} so that the operators can take  actions to investigate and resolve the underlying issues. The ubiquitous use of networked sensors and actuators in CPSs and other systems (e.g. autonomous vehicles) will become even more prevalent with the emergence of the Internet of Things (IoT), leading to multiple systems and devices communicating and possibly operating a large variety of tasks autonomously over networks.  As many of the CPSs are engineered for mission-critical tasks, they are prime targets for cyber-attacks.  It is thus of particular importance to closely monitor the behaviors of these systems for intrusion events through anomaly detection using the multivariate time series data generated by the systems.

An anomaly is usually defined as points in certain time steps where the system's behaviour is significantly different from the previous normal status \cite{Chandola:2008}.  The basic task of anomaly detection is thus to identify the time steps in which an anomaly may have occurred.  
Traditionally, Statistical Process Control (SPC) methods such as CUSUM, EWMA and Shewhart charts were popular solutions for monitoring quality of industrial processes to find out working states that are out of range \cite{Sun:2014}. These conventional detection techniques are unable to deal with the multivariate data streams generated by the increasingly dynamic and complex nature of modern CPSs.  As such, researchers have moved beyond specification or signature-based techniques and begun to exploit machine learning techniques to exploit the large amounts of data generated by the systems\cite{Donghwoon2017}.  Due to the inherent lack of labeled data, anomaly detection is typically treated as an unsupervised machine learning task.  However, most  existing unsupervised methods are built through linear projection and transformation that is unable to handle non-linearity in the hidden inherent correlations of the multivariate time series. Also, most of the current techniques employ simple comparisons between the present states and the predicted normal ranges to detect anomalies, which can be inadequate given the highly dynamic nature of the systems. 

Recently, the Generative Adversarial Networks (GAN) framework has been proposed to build generative deep learning models via adversarial training \cite{Goodfellow2014}.  While GAN has been shown to be wildly successful in image processing tasks such as generating realistic-looking images, there has been limited work in adopting the GAN framework for time-series data todate. To the best of our knowledge, there are only few preliminary works that used GAN to generate continuous valued sequences in the literature.  Yet in these early works, the GAN framework has been proven to be effective in generating time series sequences, either to produce polyphonic music with recurrent neural networks as generator and discriminator \cite{Mogren2016}, or to generate real-valued medical time series using an conditional version of recurrent GAN \cite{Esteban2017}. These early successes of GAN in generating realistic complex datasets, as well as the simultaneous training of both a generator and a discriminator in an adversarial fashion, are highly suggestive of the use of the GAN framework for anomaly detection. 

In this work, we propose a novel Multivariate Anomaly Detection strategy with GAN (MAD-GAN) to model the complex multivariate correlations among the multiple data streams to detect anomalies using both the GAN-trained generator and discriminator.  Unlike traditional classification methods, the GAN-trained discriminator learns to detect fake data from real data in an unsupervised fashion, making it an attractive unsupervised machine learning technique for anomaly detection \cite{XueY2017}. Inspired by \cite{Raymond2016} and \cite{Salimans2016} that updates a mapping from the real-time space to a certain latent space to enhance the training of generator and discriminator, researchers have recently proposed to train a latent space understandable GAN and apply it for unsupervised learning of rich feature representations for arbitrary data distributions. \cite{Schlegl2017} and \cite{Zenati2018} showed the possibility of recognizing anomalies with reconstructed testing samples from latent space, and successfully applied the proposed GAN-based detection strategy to discover unexpected markers for images. In this work, we will leverage on these previous works to make use of both the GAN-trained generator and discriminator to detect anomalies based on both reconstruction and discrimination losses.

The rest of this paper is organized as follows. Section \ref{sec:RW} presents an overview of the related works. Section \ref{sec:Model} introduces our proposed MAD-GAN framework and the anomaly score function. In Section \ref{sec:ConcandFW}, we introduce the tested CPSs and datasets. In Section \ref{sec:ExpandRes}, we show the experimental results of our proposed MAD-GAN on two real-world CPS datasets. Finally, Section \ref{sec:ConcandFW} summarizes the whole paper and suggests possible future work.

\section{Related Works}
\label{sec:RW}

Given the inherent lack of labeled anomaly data for training supervised algorithms,  anomaly detection methods are mostly based on unsupervised methods. We can divide the unsupervised detection methods into four categories: i) linear model-based method, ii) distance-based methods, iii) probabilistic and density estimation-based methods, and vi) the deep learning-based methods that are recently highly popular.

For linear model-based unsupervised anomaly detection methods, a popular approach is the Principal Component Analysis (PCA) \cite{Li:2014}. PCA is basically a multivariate data analysis method that preserves the significant variability information extracted from the process measurements and reduces the dimension for huge amount of correlated data \cite{Wold1987}. PLS is another multivariate data analysis method that has been extensively utilized for model building and anomaly detection \cite{Herman1985}. However, they are only effective for highly correlated data, and require the data to follow multivariate Gaussian distribution \cite{XuewuD2013}. 

For the distance-based methods, a popular approach is the K-Nearest Neighbor (KNN) algorithm which computes the average distance to its k nearest neighbours and obtains anomaly scores based on this distane\cite{Fabrizio:2002}.  The Clustering-Based Local Outlier Factor (CBLOF) method is another example of the distance-based methods.  It uses a predefined anomaly score function to identify anomalies based on clustering, which is an enhanced version of the Local Outlier Factor (LOF) method \cite{Breunig:2002}. Although effective in some cases, these distance-based methods perform better with priori knowledge about anomaly durations and the number of anomalies.

The probabilistic model-based and density estimation-based methods were proposed as improvements of distance-based methods by paying more attention to the data distributions. For example, the Angle-Based Outlier Detection (ABOD) method \cite{Kriegel:2008} and Feature Bagging (FB) method \cite{Lazarevic:2005} deal with data by taking variable correlations into consideration. However, these methods are unable to take into consideration the temporal correlation along time steps, and thus do not work well for multivariate time series data.

The deep learning-based unsupervised anomaly detection methods have gained much popularity recently with their promising performance. For instance, the Auto-Encoder (AE) \cite{Jiawei2011} is a popular deep learning model for anomaly detection by inspecting its reconstruction errors. Others like Deep Autoencoding Gaussian Mixture Model (DAGMM) \cite{ZongBo:2018} and LSTM Encoder-Decoder \cite{Habler:2018} have also reported good performance for multivariate anomaly detection. In this work, we follow the  promising success of deep learning-based unsupervised anomaly detection methods, and propose a novel deep learning-based unsupervised anomaly detection strategy  built on the basis of the Generative Adversarial Networks (GAN). 

Our contributions of this paper are summarized as follows: (i) we proposed MAD-GAN, a GAN-based unsupervised anomaly detection method  to detect anomalies for multivariate time series; (ii) the MAD-GAN architecture adapts the GAN framework previously developed for  image-related applications to analyze multivariate time series data by adopting the Long Short Term-Recurrent Neural Networks (LSTM-RNN) as the base models learned by the GAN to capture the temporal dependency; (iii) we used both GAN's discriminator and generator to detect anomalies using a novel anomaly score that combines the discrimination results and reconstruction residuals for each testing sample.  The proposed MAD-GAN is shown to outperform existing methods in detecting anomalies caused by cyber-attacks for two CPS datasets.

\section{Anomaly Detection with Generative Adversarial Training}
\label{sec:Model}
\begin{figure*}[!t]
\centering
\includegraphics[width=1.0\textwidth]{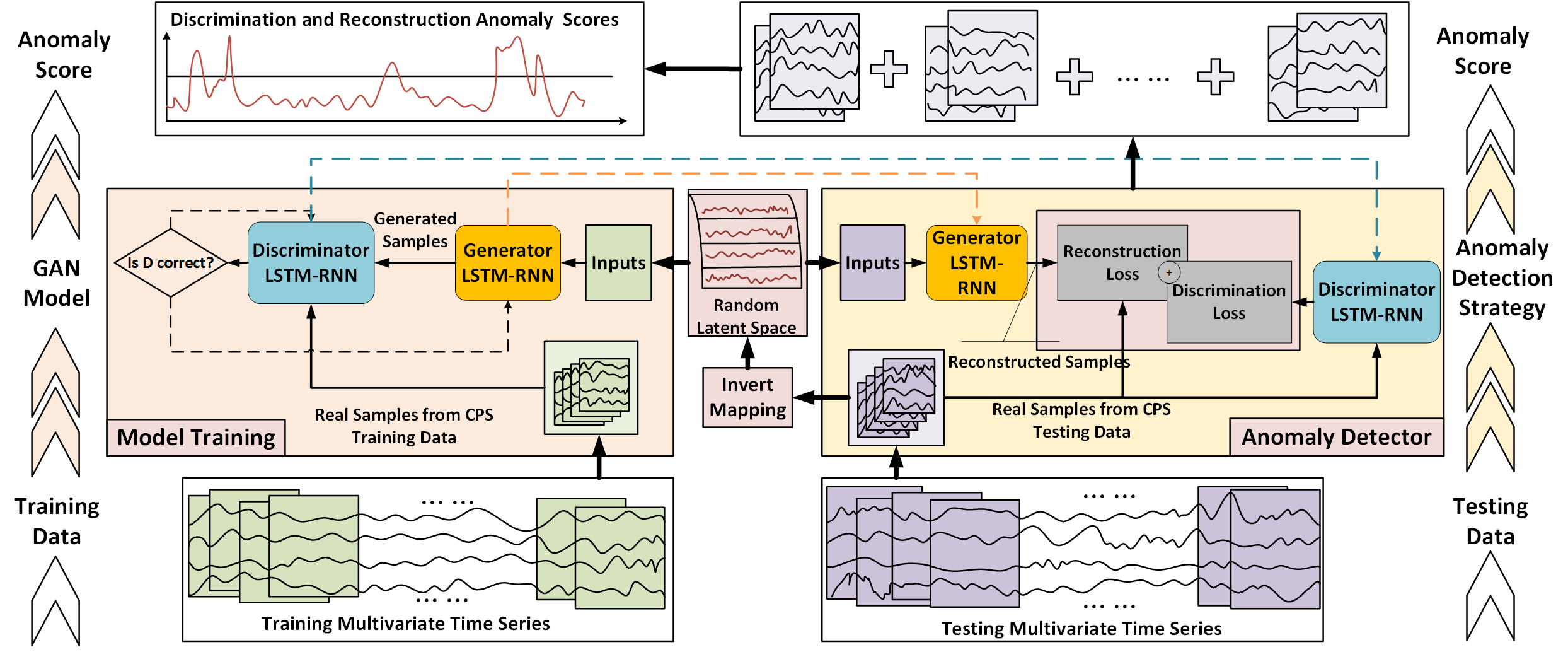}
\vspace{-0.5cm}
\caption{MAD-GAN: Unsupervised GAN-based anomaly detection. On the left is a GAN framework in which the generator and discriminator are obtained with iterative adversarial training. On the right is the anomaly detection process where both the GAN-trained discriminator and generator are applied to compute a combined anomaly score based on discrimination and reconstruction.}
\label{fig:MAD-GAN}
\end{figure*}

The basic task of anomaly detection for time series is to identify whether the testing data conform to the normal data distributions; the non-conforming points are called anomalies, outliers, intrusions, failures or contaminants in various application domains \cite{Donghwoon2017}. Fig. \ref{fig:MAD-GAN} depicts the overall architecture of the proposed MAD-GAN.

\subsection{MAD-GAN Architecture}
\label{subsec:MAD-GAN}
 First, to handle the time-series data, we construct the GAN's generator and discriminator as two Long-Short-Term Recurrent Neural Networks (LSTM-RNN), as shown in the left middle part of Fig. \ref{fig:MAD-GAN}. Following a typical GAN framework, the generator ($G$) generates fake time series with sequences from a random latent space as its inputs, and passes the generated sequence samples to the discriminator ($D$), which will try to distinguish the generated (i.e. ``fake'') data sequences from the actual (i.e. ``real'') normal training data sequences.  
 
Instead of treating each data stream independently, the MAD-GAN framework considers the entire variable set concurrently in order to capture the latent interactions amongst the variables into the models.  We divide the multivariate time series into sub-sequences with a sliding window before discrimination. To empirically determine an optimal window length for sub-sequences representation, we used different window sizes to capture the system status at different resolutions, namely $s_w=30\times i, i=1,2,...,10$. 

As in standard GAN framework, the parameters of $D$ and $G$ are updated based on the outputs of $D$, so that the discriminator can be trained to be as sensitive as possible to assign correct labels to both real and fake sequences, while the generator will be trained to be as smart as possible to fool the discriminator (i.e. to mislead $D$ to assign real labels to fake sequences) after sufficient rounds of iterations. By being able to generate realistic samples, the generator $G$ will have captured the hidden multivariate distributions of the training sequences and can be viewed as an implicit model of the system at normal status. At the same time, the resulting discriminator $D$ has also been trained to be able to distinguish fake (i.e. abnormal) data from real (i.e. normal) data with high sensitivity. In this work, we propose to exploit both $G$ and $D$ for the anomaly detection task by (i) reconstruction: exploiting the residuals between real-time testing samples and reconstructed samples by $G$ based on the mapping from real-time space to the GAN latent space; and (ii) discrimination: using the discriminator $D$ to classify the time series. This is depicted in the right middle part of Fig. \ref{fig:MAD-GAN}. As shown, the testing samples are mapped back into the latent space to calculate the corresponding reconstruction loss  based on the difference between the reconstructed testing samples (by the generator $G$) and the actual testing samples. At the same time, the testing samples are also fed to the trained discriminator $D$ to compute the discrimination loss.  Note that the testing multivariate time series are similarly divided into a set of sub-sequences by sliding window and before being fed into the detection model. We use a novel Discrimination and Reconstruction Anomaly Score (DR-Score) to combine the two losses to detect potential anomalies in the data (more details are described in Section \ref{subsec:DRS}).

\subsection{GAN-based Anomaly Detection}
\label{subsubsec:Prob}
Let us now formulate the anomaly detection problem using GAN. Given a training dataset $\mathcal{X}\subseteq \mathcal{R}^{M\times T}$ with $T$ streams and $M$ measurements for each stream, and a testing dataset $\mathcal{X}^{test}\subseteq \mathcal{R}^{N\times T}$ with $T$ streams and $N$ measurements for each stream, the task is to assign binary ($0$ for normal and $1$ for anomalous) labels to the measurements of testing dataset. Note that we assume here that all the points in the training dataset are normal.  

To effectively learn from $X$,  we apply a sliding window with window size $s_w$ and step size $s_s$ to divide the multivariate time series into a set of multivariate sub-sequences $X=\lbrace x_i, i=1,2,...,m \rbrace \subseteq \mathcal{R}^{s_w\times T}$, where $m=\frac{(M-s_w)}{s_s}$ is the number of sub-sequences. Similarly, $Z=\lbrace z_i, i=1,2,...,m\rbrace$ is a set of multivariate sub-sequences taken from a random space. By feeding $X$ and $Z$ to the GAN model, we train the generator and discriminator  with the following two-player minimax game:
\begin{equation}
\begin{array}{lcl}
\label{eq:gan_minmax}
\min \limits_G \max \limits_D V(D,G) = \mathcal{E}_{x\sim p_{data}(X)}\left[ \log D(x)\right]\\
\;\;\;\;\;\;\;\;\;\;\;\;\;\;\;\;\;\;\;\;\;\; + \mathcal{E}_{z\sim p_z(Z)}\left[\log (1-D(G(z)))\right]
\end{array}
\end{equation}
 
In this work, both the generator ($G$) and discriminator ($D$) of GAN are Long Short Term-Recurrent Neural Networks (LSTM-RNN).  After sufficient rounds of training iterations, the trained discriminator $D_{rnn}$ and the generator $G_{rnn}$ can then be employed to detect anomalies in $\mathcal{X}^{test}$ using a combined Discrimination and Reconstruction Anomaly Score (DR-Score), which will be introduced in Section \ref{subsec:DRS}. 

For detection, the testing dataset $\mathcal{X}^{test}\subseteq \mathcal{R}^{N\times T}$ is similarly divided into multivariate sub-sequences $X^{tes}={x^{tes}_j, j=1,2,...,n}$ with a sliding window, where $n=\frac{(N-s_w)}{s_s}$. Using the computed DR-Score (DRS) of the testing dataset, we label each of the sub-sequences in the testing dataset as follows:
\begin{equation}
\label{eq:mark_vec}
\begin{array}{lll}
A_t^{tes}=\left\{\begin{matrix}
1, & if\; H\left(DRS_t,1\right)>\tau\\
0, & else
\end{matrix}\right.
\end{array}
\end{equation}
where $A_t^{tes}\subseteq \mathcal{R}^{N\times 1}$ is a label vector for the testing dataset, where non-zero values indicate an anomaly is detected, i.e. the cross entropy error $H\left(.,.\right)$ for the anomaly score is higher than a predefined value $\tau$.

\begin{algorithm}[t!]
\begin{algorithmic}
  \LOOP 
   \IF {epoch within number of training iterations}
    \FOR {the $k^{th}$ epoch}
     \STATE Generate samples from the random space: 
     \STATE $Z=\lbrace z_i, i=1,...,m\rbrace \Rightarrow G_{rnn}(Z)$
     \STATE Conduct discrimination: 
     \STATE $X=\lbrace x_i, i=1,...,m\rbrace \Rightarrow D_{rnn}(X) $  
     \STATE $G_{rnn}(Z) \Rightarrow D_{rnn}(G_{rnn}(Z))$
     \STATE Update discriminator parameters by minimizing(descending) $D_{loss}$:
        \STATE $\min \frac{1}{m}\sum_{i=1}^{m}\left[-\log D_{rnn}(x_i)-\log (1-D_{rnn}(G_{rnn}(z_i)))\right]$
     \STATE Update discriminator parameters by minimizing(descending) $G_loss$ :
        \STATE $\min \sum_{i=1}^{m}\log (-D_{rnn}(G_{rnn}(z_i)))$
     \STATE Record parameters of the discriminator and generator in the current iteration.    
    \ENDFOR   
   \ENDIF
   \FOR {the lth iteration}
    \STATE Mapping testing data back to latent space:
       \STATE $Z^k=\min\limits_{Z} Er(X^{tes}, G_{rnn}(Z^i))$
   \ENDFOR
   \STATE Calculate the residuals:
       \STATE $Res = \mid X^{tes}- G_{rnn}(Z^k)\mid$
   \STATE Calculate the discrimination results:
       \STATE $Dis=D_{rnn}(X^{tes})$
   \STATE Obtain the combined anomaly score:
       \FOR {k,j and s in ranges}
           \IF{j+s=k}
             \STATE $R=\lambda Res +(1-\lambda)Dis$
             \STATE $DRS_k=\frac{\sum R_{j,s}}{L_k}$
            \ENDIF
       \ENDFOR
  \ENDLOOP
\end{algorithmic}
\caption{LSTM-RNN-GAN-based Anomaly Detection Strategy}
\label{algo:lstm-rnn-gan}
\end{algorithm}

\subsection{DR-Score: Anomaly Detection using both Discrimination and Reconstruction}
\label{subsec:DRS}

An advantage of using GAN is that we will have a discriminator and a generator trained simultaneously.  We propose to exploit both the discriminator and generator that has been jointly trained to represent the normal anatomical variability for identifying anomalies. Following the formulation in \cite{Schlegl2017}, the GAN-based anomaly detection consists of the following two parts:
\begin{enumerate}
\item \textbf{Discrimination-based Anomaly Detection }\\
Given that the trained discriminator $D$ can distinguish fake data (i.e. anomalies) from real data with high sensitivity, it serves as a direct tool for anomaly detection.
\item \textbf{Reconstruction-based Anomaly Detection }\\
The trained generator $G$, which is capable of generating realistic samples, is actually a mapping from the latent space to real data space: $G(Z):Z\rightarrow X$, and can be viewed as an inexplicit system model that reflects the normal data's distribution. Due to the smooth transitions of latent space mentioned in \cite{Radford2015}, the generator outputs similar samples if the inputs in the latent space are close. Thus, if it is possible to find the corresponding $Z^k$ in the latent space for the testing data $X^{tes}$, the similarity between $X^{tes}$ and $G(Z^k)$ (which is the reconstructed testing samples) could explain to which extent is $X^{tes}$ follows the distribution reflected by $G$. In other words, we can also use the residuals between $X^{tes}$ and $G(Z^k)$  for identifying anomalies in testing data.\\
\end{enumerate}

To find the optimal $Z^k$ that corresponds to the testing samples, we first sample a random set $Z^1$ from the latent space and obtain reconstructed raw samples $G(Z^1)$ by feeding it to the generator (as shown in the right part of Fig. \ref{fig:MAD-GAN}). Then, we update the samples from the latent space   with the gradients obtained from the error function defined with $X^{tes}$ and $G(Z)$.
\begin{equation}
\label{eq:invLoss}
\min\limits_{Z^k} Er(X^{tes}, G_{rnn}(Z^k))=1-Simi(X^{tes},G_{rnn}(Z^k))
\end{equation} 
where the similarity between sequences could be defined as covariance for simplicity.

After enough iteration rounds such that the error  is small enough, the samples $Z^k$ is recorded as the corresponding mapping in the latent space for the testing samples. The residual at time $t$ for testing samples is calculated as
\begin{equation}
\label{eq:residual}
Res(X^{tes}_t)=\sum_{i=1}^n \mid x_t^{tes,i}-G_{rnn}(Z^{k,i}_t) \mid 
\end{equation}
where $X^{tes}_t\subseteq \mathcal{R}^n$ is the measurements at time step $t$ for $n$ variables. In other words, the the anomaly detection loss is
\begin{equation}
\label{eq:score}
L_t^{tes}=\lambda Res(X^{tes}_t)+(1-\lambda)D_{rnn}(X^{tes}_t)
\end{equation}

Based on the above descriptions, the GAN-trained discriminator and generator will output a set of anomaly detection losses $\{L=L_{j,s},j=1,2,...,n;s=1,2,...,s_w\}\subseteq \mathcal{R}^{n\times s_w}$ for each test data sub-sequence. We compute a combined discrimination-cum-reconstruction anomaly score called the DR-Score (DRS) by mapping the anomaly detection loss of sub-sequences back to the original time series:
\begin{equation}
\label{eq:DRS}
\begin{array}{lll}
DRS_t = \frac{\sum \limits_{j,s\in \{j+s=t\}} L_{j,s}}{lc_t} \\
lc_t = count(j,s\in \{j+s=t\})
\end{array}
\end{equation}
where $t\in \{1,2,...,N\}$, $j\in \{1,2,...,n\}$, and $s\in \{1,2,..., s_w\}$. 

The proposed MAD-GAN is summarized in Algo. \ref{algo:lstm-rnn-gan}. We used mini-batch stochastic optimization based on Adam Optimizer and Gradient Descent Optimizer for updating the model parameters in this work.

\section{CPSs and Cyber-attacks}
\label{sec:CPSandAtt}

\subsection{Water Treatment and Distribution System}
\label{subsec:CPS}

\subsubsection{SWaT}
The Secure Water Treatment (SWaT) system is an operational test-bed for water treatment that represents a small-scale version of a large modern water treatment plant found in large cities \cite{Mathur2016}. The overall testbed design was coordinated with Singapore's Public Utility Board, the nation-wide water utility company, to ensure that the overall physical process and control system closely resemble real systems in the field. The SWaT dataset collection process lasted for a $11$ days with the system operated 24 hours per day. A total of $36$ attacks were launched during the last $4$ days of 2016 SWaT data collection process \cite{Jonathan2016}. Generally, the attacked points include sensors (e.g., water level sensors, flow-rate meter, etc.) and actuators (e.g., valve, pump, etc.). These attacks were launched on the test-bed with different intents and diverse lasting durations (from a few minutes to an hour) in the final four days. The system was either allowed to reach its normal operating state before another attack was launched or the attacks were launched consecutively. Please refer to the SWat website\footnote{https://itrust.sutd.edu.sg/testbeds/secure-water-treatment-swat/} for more details about the SWaT dataset. 

The water purification process in SWaT is composed of six sub-processes referred to as $P1$ through $P6$ \cite{Jonathan2016}. The first process is for raw water supply and storage, and $P2$ is for pre-treatment where the water quality is assessed. Undesired materials are them removed by ultra-filtration (UF) backwash in $P3$. The remaining chorine is destroyed in the Dechlorination process ($P4$). Subsequently, the water from $P4$ is pumped into the Reverse Osmosis (RO) system ($P5$) to reduce inorganic impurities. Finally, $P6$ stores the water ready for distribution. 

\subsubsection{WADI}
Unlike a water treatment system plant which is typically contained in a secured location, a distribution system comprises numerous pipelines spanning across a large area. This highly increases the risk of physical attacks on a distribution network. The Water Distribution (WADI) testbed is  an extension of the SWaT system, by taking in a portion of SWaT’s reverse osmosis permeate and raw water to form a complete and realistic water treatment, storage and distribution network. There are three control processes in the water distribution system. The first process is to intake the raw water from SWaT, Public Utility Board (PUB) inlet or the return water in WADI, and store the raw water in two tanks. $P2$ distributes water from two elevated reservoir tanks and six consumer tanks based on a pre-set demand pattern. Water is recycled and sent back to $P1$ at the third process.

The WADI testbed is similarly equipped with chemical dosing systems, booster pumps and valves, instrumentation and analysers \cite{AhmedCM2017}.  In addition to simulating attacks and defences being carried out on the PLCs via networks, WADI has the capabilities to simulate the effects of physical attacks such as water leakage and malicious chemical injections. The WADI data collection process consists of 16 days' continuous operations, of which $14$ days’ data was collected under normal operation and $2$ days' with attack scenarios. During the data collection, all network traffic, sensor and actuator data were collected. Please refer to the WADI website\footnote{https://itrust.sutd.edu.sg/testbeds/water-distribution-wadi/} for more details about the WADI dataset.

\subsection{Cyber-Attacks}
\label{subsec:Atta}

The goal of an attacker is to manipulate the normal operations of the plant. It is assumed that the attacker has remote access to the SCADA system of SWaT and WADI and has the general knowledge about how the systems work. Various experiments have been conducted on the SWaT and WADI systems to investigate cyber-attacks and respective system responses. In total, 36 attacks and 15 attacks have been inserted to SWaT and WADI, respectively \cite{Mathur2016,MathurCM2017}. For illustration, let us explain one exemplary attack for each system.
\begin{itemize}
\item \textbf{SWaT}
One attacking goal is to degrade the performance of SWaT from the nominal level (for example, 5 gallons/minute) to a lower value. This attack could be launched by compromising sensor LIT401, which measures the water level of the Reverse Osmosis (RO) feed tank in $P4$. By attacking LIT401, the attacker reduced the level of the RO feed tank from 800mm to 200mm, which would lead the PLC-4 to stop the pump P401 and less water was pumped to $P5$. Finally he negative impact of attacking sensor LIT401 was reflected on the output water flow rate of RO unit (values measured by FIT501 in $P5$). According to system specifications, this flow rate must remain at about 1.2cm/hr which leads to nearly 5 gallons/minute of treated water. Thus, the amount of treated water was reduced during the observation period.
\item \textbf{WADI}
One attacking goal is to tamper the readings of water level sensor in $P1$. The attacker altered the sensor reading from $76\%$ to $10\%$ of the tank capacity which indicated a ``low state''. As a result, the PLC-1 (controller of $P1$) sent a command to turn on the intake water pump to intake more water from WADI return, SWaT output or PUB inlet. At the same time, due to the false low water level state in the raw water tank, the water supply from $P1$ to $P2$ was cut off while $P2$ continued to supply water to the consumer tanks. Thus, water levels of tanks in $P2$ decreased. Once the water level in the elevated tanks ($P2$) reached a low level, the supply to consumer tanks ($P2$) was cut off. Consequently, by tampering the readings of water level sensor in $P1$ to a low level, there would be an overflow in the tanks of $P1$ and no water flow in $P2$.
\end{itemize}

\begingroup
\setlength{\tabcolsep}{10pt} 
\begin{table}[h]
  \caption{General Information about Datasets}
  \label{tab:datasets}
  \centering
  \newcommand{\cc}[1]{\multicolumn{1}{c}{#1}}
  \begin{tabular}{llll}
  \hline\hline
    \toprule
          Item& SWaT& WADI& KDDCUP99 \\
    \midrule
          Variables& 51& 103& 34\\
          Attacks& 36& 15& 2\\
          Attack durations(mins)& 2$\sim$ 25& 1.5$\sim$ 30 &NA\\
          Training size (normal data)& 496800& 1048571& 562387 \\
          Testing size (data with attacks)& 449919& 172801& 494021\\
          N\_rate(\%)& 88.02& 94.01& 19.69\\
  \hline\hline
  \multicolumn{3}{l}{*N\_rate is the rate of normal data points versus all in testing data sets. }\\
  \hline
  \bottomrule
  \end{tabular}
\end{table}
\endgroup

\subsection{SWaT and WADI Datasets}
\label{subsec:Dataset}
The SWaT/WADI data collection process lasted for a 11/16 days with the systems operated 24 hours per day. Various cyber-attacks were executed on the test-beds with different intents and divergent lasting durations (from a few minutes to an hour) in the final 4/2 days. The systems were either allowed to reach its normal operating state before another attack was launched or the attacks were launched consecutively. Some general information about these two datasets are summarized in Table \ref{tab:datasets}. To better understand the complexity of the detection tasks, it is worthwhile to note that the two datasets and their associated normal conditions and attacks have the following characteristics:
\begin{itemize}
\item Different attacks may last for different time durations due to different scenarios. Some attacks did not even take effect immediately. The system stabilization durations also vary across attacks. Simpler attacks, such as those aiming at changing flow rates, require less time for the system to stabilize while the attacks that caused stronger effects on the dynamics of system  will require more time for stabilization. 
\item Attacks on one sensor/actuator may affect the performance on other sensors/actuators or the whole system (this point can be seen from the two examples in Section \ref{subsec:Atta}), usually after a certain time delay.
\item Furthermore, similar types of sensors/actuators tend to respond to attacks in a similar fashion. For example, attacks on the LIT101 sensor (a water level sensor in $P1$ of SWaT) caused obvious abnormal spikes in both LIT101 and LIT301 (another water level sensor in $P3$ of SWaT) but the affects on readings of other sensors and actuators (such as flow rate sensors and power meters) were relatively smaller.
\end{itemize}

The aforementioned observations suggest that it is of importance to adopt a multivariate approach in the modeling of the systems for anomaly detection, as the overall change in performance by different sub-process could collectively help to better recognize the attacks. In other words, the underlying correlations between the sensors and actuators could be useful to detect the anomalies in the behaviors of the system caused by the attack.

\section{Experiments}
\label{sec:ExpandRes}

\subsection{Data Preparation and System Architecture}
\label{subsec:DataandSys}

In the SWaT dataset, $51$ variables (sensor readings and actuator states) were measured for 11-days. Within the raw data, $496,800$ samples were collected under normal working conditions (data collected in the first 7-days), and $449,919$ samples were collected when various cyber-attacks were inserted to the system subsequently. Similarly, for the WADI dataset, $789371$ samples for $103$ variables were collected under normal working conditions in the first 14-days, and $172801$ samples were collected when various cyber-attacks were inserted to the system in the last 2 days. For both of these datasets, we eliminated the first $21,600$ samples from the training data (normal data) since it took 5-6 hours to reach stabilization when the system was first turned on according to \cite{Jonathan2016}.

In the anomaly detection process, we subdivide the original long multiple sequences into smaller time series by taking a sliding window across raw streams. Since it is an important topic in time series study to decide the optimal window length for sub-sequences representation, we tried a set of different window sizes to capture the system status at different resolutions, namely $s_w=30\times i, i=1,2,...,10$. To capture the relevant dynamics of SWaT data, the window is applied to the normal and testing datasets with shift length $s_s$=$10$. 

For this study,  we used an LSTM network with depth 3 and 100 hidden (internal) units for the generator. The LSTM network for the discriminator is relatively simpler with 100 hidden units and depth 1. Inspired by the discussion about latent space dimension in \cite{Esteban2017}, we also tried different dimensions and found that  higher latent space dimension generally generates better samples especially when generating multivariate sequences. We set the dimension of latent space as $15$ in this study.

\subsection{Evaluation Metrics}
\label{subsec:EvMe}

We use the standard metrics, namely Precision (Pre), Recall (Rec), and F\_1 scores, to evaluate the anomaly detection performance of MAD-GAN:
\begin{equation}
\label{eq:Pre}
Pre=\frac{TP}{TP+FP}
\end{equation}
\begin{equation}
\label{eq:Rec}
Rec=\frac{TP}{TP+FN}
\end{equation}
\begin{equation}
\label{eq:F1}
F_{1}=2\times \frac{Pre \times Rec}{Pre+Rec}
\end{equation}
where $TP$ is the correctly detected anomaly (True Positives: $A_t=1$ while real label $L_t=1$), $FP$ is the falsely detected anomaly (False Positives: $A_t=1$ while real label $L_t=0$), $TN$ is the correctly assigned normal (True Negatives: $A_t=0$ while real label $L_t=0$), and $FN$ is the falsely assigned normal (False Negatives: $A_t=0$ while real label $L_t=1$). 


Given that our application in this work is to detect intrusions (cyber-attack), it is important for the system to detect all the attacks even if it requires tolerating a few false alarms.  As such, the focus is on correctly detecting  actual positives,  while the false positives are not so important as long as they are not excessive. Therefore, we use recall as the main metric to measure the performance of anomaly detection for this study.

\subsection{Results}
\label{subsec:ADRes}
We evaluate the anomaly detection performance of MAD-GAN on the aforementioned two datasets SWaT and WADI. As described earlier, the sub-sequences are fed into the MAD-GAN model.  Note that to reduce the computational load, we reduce the original dimension by PCA, choosing the PC dimension in based on the PC variance rate. For comparison on the anomaly detection performance, we also applied PCA, K-Nearest Neighbour (KNN), Feature Bagging (FB), and Auto-Encoder (AE) that are popular unsupervised anomaly detection methods on the datasets. To compare with a GA-based method, we  compared  MAD-GAN with the Efficient GAN-based (EGAN) method of \cite{Zenati2018} whose discriminator and generator were implemented as fully connected neural networks. 

\subsubsection{Multivariate Generation}
\label{subsubsec:MultiGen}
\begin{figure}[!h]
\centering
\begin{minipage}{1.0\textwidth}
  \centering
\includegraphics[width=1.0\textwidth, height=4.5cm]{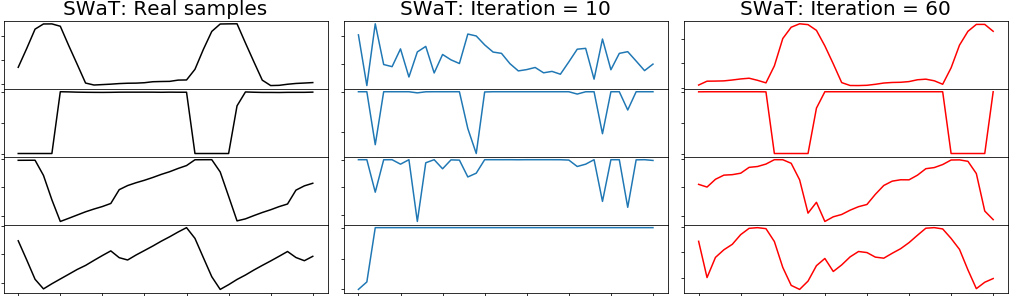}
\end{minipage}\\
\begin{minipage}{1.0\textwidth}
  \centering
\includegraphics[width=1.0\textwidth, height=4.5cm]{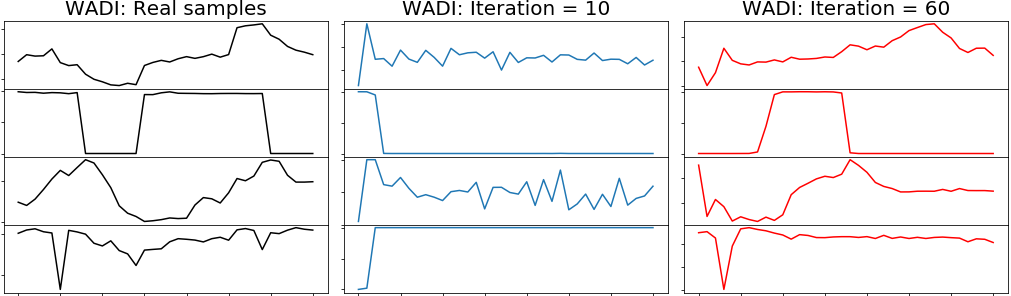}
\end{minipage}%

\caption{Comparison between generated samples at different traning stages: GAN-generated samples at early stage are quite random while those generated at later stages almost perfectly took the distribution of original samples. Note that we only plot four variables for each dataset 
as visualization examples.}
\label{fig:rs_gs}
\end{figure}

First, we visualize the multivariate data samples generated by  MAD-GAN versus the actual samples from the CPS. As can be observed in Fig. \ref{fig:rs_gs}, our GAN-generated samples that were clearly different from the training data in the early learning stage (iteration=10). However, after sufficient number of iterations, the generator was able to output realistic multivariate samples for the various sensors and actuators of the systems. Note that no PCA projection was applied to the trainings samples (real samples) in this visualization example.

In addition, we use Maximum Mean Discrepancy (MMD) \cite{Gretton2007} to evaluate whether the GAN model has learned the distributions of the training data.  MMD  is one of the training objectives for moment matching networks \cite{Yujia2015,Chun-Liang2017}. 
\begin{equation}
\label{eq:mmd}
\begin{array}{lll}
MMD(Z_j,X_{tes})=\frac{1}{n(n-1)}\sum_{i=1}^n \sum_{j\neq i}^n K(Z_i^k, Z_j^k)\\
\;\;\;\;\;\;\;\;\;\;\;\;\;\;\;\;\;\;\;\;\;\;-\frac{2}{mn}\sum_{i=1}^n \sum_{j=1}^m K(Z_i^k, X_j^{tes})\\
\;\;\;\;\;\;\;\;\;\;\;\;\;\;\;\;\;\;\;\;\;\;+\frac{1}{m(m-1)}\sum_{i=1}^m \sum_{j\neq i}^m K(X_i^{tes}, X_j^{tes})\\
\end{array}
\end{equation}

\begin{figure}[!h]
\centering
\includegraphics[width=1.0\textwidth, height=4.5cm]{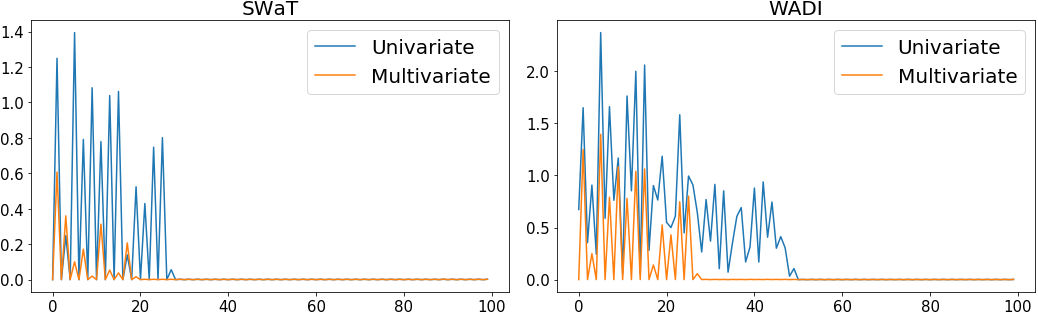}
\caption{MMD: generation for multiple time series v.s. generation single time series.} 
\label{fig:mmd}
\end{figure}

We plot the MMD values across GAN training iterations for generating  multivariate samples for both datasets in Fig. \ref{fig:mmd}. We can observe that the MMD values tend to converge to small values after 30-50 iterations for both datasets. We also compared the MMD values for univaraite sample generation. Interestingly,  the early MMD values of multivariate samples were lower than that of univariate samples, and the MMD for multivariate samples also converged faster than the univariate case. This suggests that using multiple data streams can help with the training of GAN model. 

\subsubsection{Anomaly Detection Performance}
\label{subsubsec:Comp}
In Table \ref{tab:detectionRes}, we show the best performance by the popular unsupervised methods (PCA, KNN, FB and AE) with underlines, and the overall  best performance in bold. As mentioned in Section \ref{subsec:DataandSys},  MAD-GAN was tested with multiple sub-sequence length resolutions. 

\begingroup
\setlength{\tabcolsep}{15pt} 
\begin{table}[!t]
	\caption{Anomaly Detection Metrics for Different Datasets}
	\label{tab:detectionRes}
	\centering
	\newcommand{\cc}[1]{\multicolumn{1}{c}{#1}}
	\begin{tabular}{lllll}
		\hline\hline
		\toprule
		Datasets& Methods& Pre& Rec& F\_1\\
		\hline
		\midrule
		\multirow{8}{*}{SWaT}&PCA&    24.92& 21.63& 0.23\\
		                     &KNN&    7.83& 7.83& 0.08\\
		                     &FB&    10.17& 10.17& 0.10\\              
		                     &AE&    \underline{72.63}& \underline{52.63}& \underline{0.61}\\ \cline{2-5}
		                     &EGAN&  40.57& 67.73& 0.51\\
		                     &MAD-GAN*& \textbf{99.99}& 54.80& 0.70\\
		                     &MAD-GAN**& 12.20& \textbf{99.98}& 0.22\\
		                     &MAD-GAN***& 98.97& 63.74& \textbf{0.77}\\
		\hline
		\multirow{8}{*}{WADI}&PCA&     \underline{39.53}& 5.63& 0.10\\
		                     &KNN&     7.76& 7.75& 0.08\\
		                     &FB&      8.60& 8.60& 0.09\\
		                     &AE&      34.35& \underline{34.35}& \underline{0.34}\\ \cline{2-5}
		                     &EGAN&    11.33& 37.84& 0.17\\
		                     &MAD-GAN*& \textbf{46.98}& 24.58& 0.32\\  
		                     &MAD-GAN**& 6.46& \textbf{99.99} &0.12\\ 
		                     &MAD-GAN***& 41.44& 33.92& \textbf{0.37}\\  
		\hline
        \multirow{8}{*}{KDDCUP99}&PCA&  60.66& 37.69& 0.47\\
                                 &KNN&   45.51& 18.98& 0.53\\
                                 &FB&    48.98& 19.36& 0.28\\
                                 &AE&    \underline{80.59}& \underline{42.36}& \underline{0.55}\\ \cline{2-5}
                                 &EGAN&    92.00& 95.82& \textbf{0.94}\\
                                 &MAD-GAN*& \textbf{94.92}& 19.14& 0.32\\
                                 &MAD-GAN**& 81.58& \textbf{96.33}& 0.88\\  
                                 &MAD-GAN***& 86.91& 94.79& 0.90\\  
		\hline
		\multicolumn{5}{l}{Rows GAN-AD* list results chosen by best Precision.}\\
        \multicolumn{5}{l}{Rows GAN-AD** list results chosen by best Recall. }\\
        \multicolumn{5}{l}{Rows GAN-AD*** list results chosen by best F\_1. }\\
		\hline
		\bottomrule 
	\end{tabular}
\end{table}
\endgroup

From Table \ref{tab:detectionRes}, we  observe the following:
\begin{itemize}
\item For the SWaT dataset, focused on the results chosen by best F\_1 since F\_1 balance precision and recall, MAD-GAN outperformed the best performance by four popular methods was given by AE by $26.34\%$ and $11.11\%$ for precision and recall, respectively. In fact, MAD-GAN achieved nearly $100\%$ precision and recall here, detecting all the anonymous points correctly for SWaT without false alarms. 
\item For the WADI dataset, also focused on the results chosen by best F\_1, recall by MAD-GAN is slightly poorer ($3.02\%$ lower) than that by AE. However, for the best recall case, MAD-GAN outperformed others by $65.64~94.36\%$ based on the recall values. Although the MAD-GAN's precision seems poor, it can achieve a near $100\%$ recall value. This is acceptable in the cyber-attack setting as the cost for false positive is tolerable for detecting all the intrusions (as mentioned in Section \ref{subsec:EvMe}). In comparison, none of the popular detection methods can achieve a satisfactory recall. 
\item Between the two  datasets,  MAD-GAN performed markedly better for SWaT. As shown by the ``N\_ rate'' of Table \ref{tab:datasets}, the WADI dataset is more unbalanced than SWaT (i.e. more actual negatives), which leads to more falsely reported positives. In addition, we also note that as shown in Table \ref{tab:datasets}, the WADI dataset has a larger feature dimension than SWaT (WADI has $103$ variables while SWaT only has $51$ variables.). 
\item Here, we also applied MAD-GAN to a more balanced dataset, the KDDCUP99 dataset. On this dataset, MAD-GAN can reach $0.90$ F\_1 score with precision larger than $85\%$ and recall higher than $94\%$. Although The EGAN results on KDDCUP99 dataset (which were reported by \cite{Zenati2018}) are better (but not overwhelmingly better) than that by MAD-GAN, MAD-GAN performed better than EGAN for both SWaT and WADI datasets (unbalanced datasets). This is because LSTM-RNN, which is used in MAD-GAN, is capable of learning complex time series better than CNNs used in EGAN. In fact, looking at the relative performance of EGAN with other non-GAN methods, we can see that GAN-based anomaly detection is unable compete other traditional methods if we do not model temporal correlation appropriately. 
\end{itemize}

Overall, MAD-GAN consistently outperformed the popular unsupervised detection methods. The only drawback is that it takes LSTM-RNN more time to deal with longer sub-sequences (to be specific, the model becomes slow when the sub-sequence length $s_w$ is larger than $200$). It will be worthwhile to explore using other Neural Networks to incorporate the temporal correlation and consider the choice of sub-sequence length for future work.

\subsubsection{Dimension Reduction}

As mentioned, to minimize the computation load of the LSTM-RNNs, we used PCA to project the raw data into a lower dimensional principal space instead of directly feeding the high dimensional data to the MAD-GAN model. We plot the variance rates of the first 10 Principal Components (PC) for both datasets  in Fig. \ref{fig:PCA} below. 

\begin{figure}[h]
\centering
\begin{minipage}{0.49\textwidth}
  \centering
\includegraphics[width=1.0\textwidth]{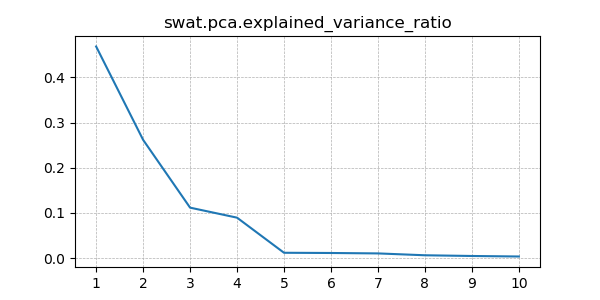}
\label{fig:swat_Acc_bp}
\end{minipage}%
\begin{minipage}{0.49\textwidth}
  \centering
\includegraphics[width=1.0\textwidth]{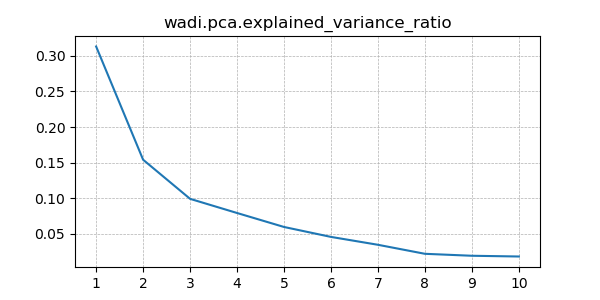}
\label{fig:swat_Pre_bp}
\end{minipage}%

\caption{Variance Ratio of Principal Component for the SWaT and WADI data.} 
\label{fig:PCA}
\end{figure}

The figure showed that one main PC explained more than $50\%$ ($30\%$) of the variance for the SWaT (WADI) data.  Also, the PCs after the $5^{th}$ ($8^{th}$) one contributed little to the overall variance (near to $0$). As such, we projected the SWaT (WADI) data to the first $5$ ($8$), and then applied the MAD-GAN to detect anomalies for the projected data. 

While the computation load was reduced largely by dimension reduction (reducing more than $\frac{1}{3}$ of the overall training-testing time) information could be lost,  which may affect the generation of realistic time series data.  To show that  the anomaly detection performance was not influenced by removing the less important variables,  in Table \ref{tab:PC_reso}, we list the anomaly detection evaluation metrics of MAD-GAN at different PC resolution (from PC=1 to PC=10) as well as all the original variable space for the SWaT dataset with sub-sequence length $s_w=30$.

\begingroup
\setlength{\tabcolsep}{5pt} 
\begin{table}[!t]
  \caption{Anomaly Detection Metrics of MAD-GAN at Different PC Resolutions}
  \label{tab:PC_reso}
  \centering
  \newcommand{\cc}[1]{\multicolumn{1}{c}{#1}}
  \begin{tabular}{llllllllllll}
  \hline\hline
    \toprule
     EM& PC=1& PC=2& PC=3& PC=4& PC=5& PC=6& PC=7& PC=8& PC=9& PC=10& ALL\\
    \hline
    \midrule
    Pre&  16.90& 17.76& 18.57& 14.53& 26.56& 24.39& 13.37& 14.78& 13.60& 13.83& 13.95\\
    Rec&  72.03& 95.34& 92.76& 91.16& 95.27& 81.25& 91.87& 92.02& 95.06& 96.03& 92.96\\      
    F\_1& 0.24&  0.23&  0.23&  0.23&  0.37& 0.22&  0.22&  0.22&  0.23&  0.23&  0.23\\ 
  \hline
  \multicolumn{12}{l}{EM: evaluation metrics. Sub-sequence length equals to 30.}\\
  \bottomrule 
\end{tabular}
\end{table}
\endgroup

We observe from Table  \ref{tab:PC_reso} the following:

\begin{itemize}
\item Although the first principal component contributed more than $50\%$ of the variance of the SWaT dataset (as shown in Fig. \ref{fig:PCA}), the MAD-GAN performance was not satisfactory (recall is less than $75\%$) with only this component. By adding the second principal component (the first two principal components together explained for more than $80\%$ of the variance of SWaT), the performance of MAD-GAN was largely improved and the performance (mainly evaluated by recall value) generally converged at a high level (larger than $90\%$) by adding more variable dimensions. 
\item By using the first five principal components (which explained for more than $99.5\%$ of the variance), we can achieve a recall higher than $95\%$. Although it was poorer than the recall reported by PC=2,10, the precision obtained by PC=5 was higher than others. This indicates that suitable feature subset/sub-combination selection is important  to reduce the false positive.
\item It is interesting to see that the F\_1 scores under different principal spaces were at similar low levels (about $0.22~0.24$). This was caused by the low precision, as large amount of false positives are reported by MAD-GAN for the unbalanced SWaT dataset. We will discuss more about the precision issues in the next subsection.
\end{itemize}

\subsubsection{Subsequence Resolution}
\label{subsubsec:AD_mul_reso}

\begin{figure}[!h]
\centering

\begin{minipage}{0.49\textwidth}
  \centering
\includegraphics[width=1.0\textwidth, height=4.5cm]{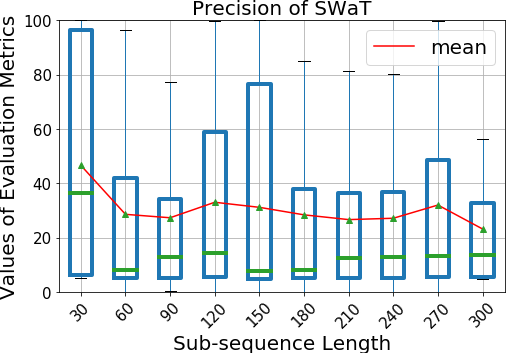}
\label{fig:swat_rec_bp}
\end{minipage}%
\begin{minipage}{0.49\textwidth}
  \centering
\includegraphics[width=1.0\textwidth, height=4.5cm]{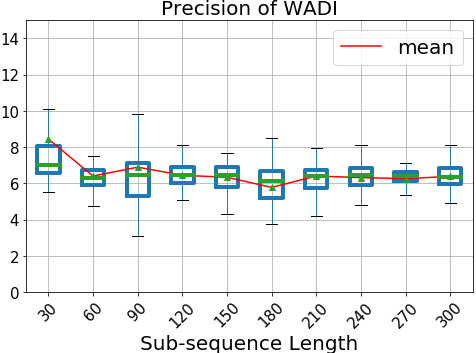}
\label{fig:wadi_rec_bp}
\end{minipage}

\begin{minipage}{0.49\textwidth}
  \centering
\includegraphics[width=1.0\textwidth, height=4.5cm]{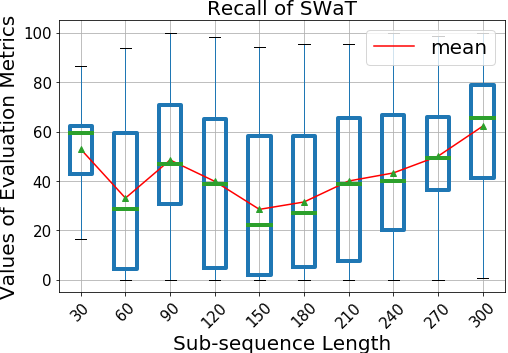}
\label{fig:swat_rec_bp}
\end{minipage}%
\begin{minipage}{0.49\textwidth}
  \centering
\includegraphics[width=1.0\textwidth, height=4.5cm]{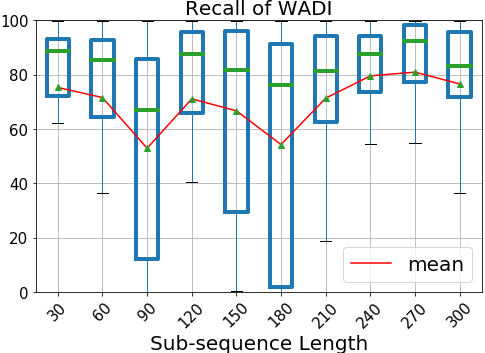}
\label{fig:wadi_rec_bp}
\end{minipage}

\begin{minipage}{0.49\textwidth}
  \centering
\includegraphics[width=1.0\textwidth, height=4.5cm]{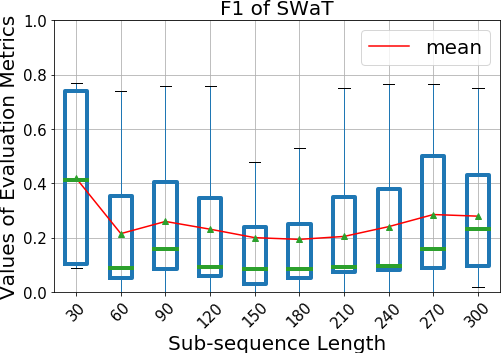}
\label{fig:swat_f1_bp}
\end{minipage}%
\begin{minipage}{0.49\textwidth}
  \centering
\includegraphics[width=1.0\textwidth, height=4.5cm]{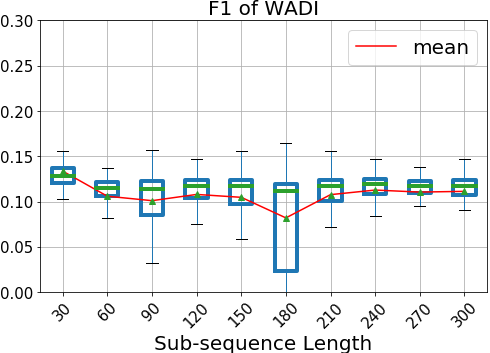}
\label{fig:wadi_f1_bp}
\end{minipage}%

\caption{Values of evaluation metrics (precision, recall and F\_1) as a function of sub-sequence lengths for SWaT and WADI. The boxes show the MAD-GAN performance at fixed 100 training iterations. The green lines in all the boxes are the median values. The mean values of each box are shown as green triangles and they are linked together with a red line.} 

\label{fig:GAN-mul-reso_swat_wadi}
\end{figure}

Selecting a suitable subsequence resolution (ie. sub-sequence length) is critical for RNN and time series related studies. In this work, we empirically determined the optimal subsequence resolution by trying a series of different sub-sequence lengths (i.e. window sizes) $s_w=30\times i,\; i=1,2,...,10$. For each sub-sequence length, the GAN model was trained recursively for $100$ iterations (i.e. epochs). We depict the box-plots of the recall and F\_1 values of MAD-GAN at each of the training iterations over different sub-sequence lengths in Figs. \ref{fig:GAN-mul-reso_swat_wadi}.  

From  Figs. \ref{fig:GAN-mul-reso_swat_wadi}, we can observe the following differences:
\begin{itemize}
\item SWat dataset: Generally, MAD-GAN achieved better metric values (precision, recall and F\_1) with SWaT dataset when the sub-sequence length used was only $3$. The precision values when using sub-sequence length $300$ were markedly poorer than those using smaller sub-sequence lengths, while the recall values were better than the average level across different sub-sequence lengths for the SWaT dataset. This indicates that there are more false positives (bad precision) with larger sub-sequence lengths for SWaT dataset.
\item WADI dataset: The overall F\_1 level for the WADI dataset was poor (less than 0.2) while the average recall values were satisfactory. This is in accordance with what we have mentioned in Section \ref{subsec:ADRes}. The poor F\_1 (as well as precision) was caused by the large amount of false positives reported for the highly unbalanced data. Nevertheless, this is acceptable given that false alarms are relatively tolerable for intrusion detection applications.
\end{itemize}

While there were fluctuations in performance  (in particular recall) and no convergence along the range of tested sub-sequence lengths, we can see that the recall values were roughly oscillating within relatively acceptable bounds ($50~100\%$) for both datasets. Thus, a relatively small sub-sequence length (such as $s_w=30$) could be a safe choice for CPS datasets like SWaT and WADI, where the data were recorded every second, and the system has relatively quick responses\footnote{It only takes several hours for SWaT and WADI to achieve steady states, and the system can recover very quickly after attacks (usually within several minutes).}. This choice of small sub-sequence length can save lots of time while the performance can be guaranteed (As mention in Section \ref{subsubsec:Comp} it took more time to train test the GAN model with sub-sequences longer than $200$). It would be a worthy  further work  for us to investigate and develop a  principled approach to determine the optimal subsequence resolution for time-series analysis using GAN.

\subsubsection{Stability across Multiple Iteration Epochs}

\begin{figure}[h]
\centering
\includegraphics[width=1.0\textwidth, height=4.5cm]{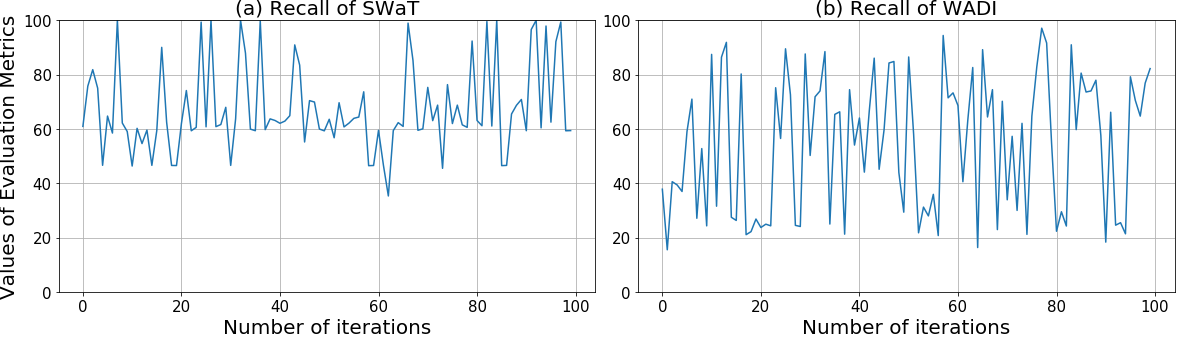}
\caption{Values of evaluation metrics as a function of the iteration epochs for the SWaT and WADI datasets when the sub-sequence length $s_w=180$.} 
\label{fig:swat_wadi_APR180_100epoch}
\end{figure}

As alluded above, the stability of MAD-GAN is another interesting issue to explore further. To observe the stability during the training epochs, we plot the recall values as a function of the iteration epochs for the SWaT and WADI datasets with sub-sequence length $s_w=180$ in in Fig. \ref{fig:swat_wadi_APR180_100epoch}. Note that the performance for 100 iterations was shown since we have fixed the number of training epochs to be $100$ for our experiments in this work. 

From the figures, we can observe  large variances of the performance (the recall values)  amongst the iteration epochs  for both datasets. On closer scrutiny, it can be seen that the performance was rapidly improved during the first few epochs and then started to oscillate at relatively high level (larger than $60\%$). We have also observed similar fluctuation when testing EGAN (its performance was also not improved by adding more training iterations).  While this  may suggest that a small number of training epochs could be sufficient for MAD-GAN to rapidly achieve satisfactory results,  the subsequent oscillations  indicate the instability of GAN-based anomaly detection method. As further work, it will be worthwhile to consider the stability of GAN-trained generators and discriminators for anomaly detection.

\section{Conclusions}
\label{sec:ConcandFW}

Today's cyber-physical systems, affixed with networked sensors and actuators, generate large amounts of data streams that can be used for monitoring the system behaviors to detect anomalies such as those caused by cyber-attacks.  In this paper, we have explored the use of GAN for multivariate anomaly detection on the time series data generated by the CPSs.  We  proposed a novel MAD-GAN (\underline{M}ultivariate \underline{A}nomaly \underline{D}etection with GAN) framework to train LSTM-RNNs on the multivariate time-series data and then utilize both the discriminator and the generator to detect anomalies using a novel Discrimination and Reconstruction Anomaly Score (DR-Score). We tested MAD-GAN on two complex cyber-attack CPS datasets from the Secure Water Treatment Testbed (SWaT) and Water Distribution System (WADI), and showed superior performance over existing unsupervised detection methods, including a GAN-based approach.

Given that this is an early attempt on multivariate anomaly detection on time series data using GAN, there are interesting issues that await further investigations.  For example, we have noted the issues of determining the optimal subsequence length as well as the potential model instability of the GAN approaches. For future work, we plan to conduct further research on feature selection for multivariate anomaly detection, and investigate principled methods for choosing the latent dimension and PC dimension with theoretical guarantees.  We also hope to perform a detailed study on the stability of the detection model.  In terms of applications, we plan to explore the use of MAD-GAN for other anomaly detection applications such as predictive maintenance and fault diagnosis for smart buildings and machineries.


%
%
%
%
%
%
%
%

\bibliographystyle{IEEEtran}
\bibliography{GANReference}

\end{document}